\documentclass[final]{elsarticle}

\usepackage{lineno,hyperref}
\usepackage{array}
\usepackage{diagbox}
\usepackage{booktabs}
\usepackage{multirow}
\usepackage[cmex10]{amsmath}
\usepackage{amsthm}
\usepackage{amssymb}
\usepackage{calligra}
\DeclareMathAlphabet{\mathcalligra}{T1}{calligra}{m}{n}
\usepackage{type1ec}
\usepackage{nicefrac}

\usepackage{algorithm}
\usepackage[noend]{algpseudocode}

\usepackage{fancyhdr}

\makeatletter
\def\BState{\State\hskip-\ALG@thistlm}
\makeatother

\usepackage[usenames,dvipsnames,table]{xcolor}

\usepackage[]{verlab}
\newcommand{\argmin}[1]{\underset{#1}{\operatorname{arg}\,\operatorname{min}}\;}

\newcommand{\fitness}[1]{\underset{#1}{\operatorname{fitness}}\;}

\journal{Journal of Visual Communication and Image Representation. }

\begin{document}

\begin{frontmatter}
	
\title{Making a long story short: A multi-importance fast-forwarding egocentric videos with the emphasis on relevant objects}



\author[address]{Michel M. Silva\corref{correspondingauthor}}
\cortext[correspondingauthor]{Corresponding author.}
\ead{michelms@dcc.ufmg.br}

\author[address]{Washington L. S. Ramos}
\ead{washington.ramos@dcc.ufmg.br}

\author[address]{Felipe C. Chamone}
\ead{cadar@dcc.ufmg.br}

\author[address]{Jo\~ao P. K. Ferreira}
\ead{joaoklock@dcc.ufmg.br}

\author[address]{Mario F. M. Campos}
\ead{mario@dcc.ufmg.br}

\author[address]{Erickson R. Nascimento}
\ead{erickson@dcc.ufmg.br}



\address[address]{Universidade Federal de Minas Gerais (UFMG), Av. Pres. Ant\^onio Carlos, 6627, Belo Horizonte, Brazil.}

\begin{abstract}
The emergence of low-cost high-quality personal wearable cameras combined with the increasing storage capacity of video-sharing websites have evoked a growing interest in first-person videos, since most videos are composed of long-running unedited streams which are usually tedious and unpleasant to watch. State-of-the-art semantic fast-forward methods currently face the challenge of providing an adequate balance between smoothness in visual flow and the emphasis on the relevant parts. In this work, we present the Multi-Importance Fast-Forward~(MIFF), a fully automatic methodology to fast-forward egocentric videos facing these challenges. The dilemma of defining what is the semantic information of a video is addressed by a learning process based on the preferences of the user.  Results show that the proposed method keeps over $3$ times more semantic content than the state-of-the-art fast-forward. Finally, we discuss the need of a particular video stabilization technique for fast-forward egocentric videos\footnote{\href{https://www.verlab.dcc.ufmg.br/semantic-hyperlapse/jvci2018/}{www.verlab.dcc.ufmg.br/semantic-hyperlapse/jvci2018/}}.

\end{abstract}


\end{frontmatter}

\thispagestyle{fancy}
\fancyhf{}
\chead{\small{In Special Issue on Egocentric Videos and Lifelogging Tools of the \\ Journal of Visual Communication and Image Representation (JVCI) 2018 \\ The final publication is available at: \href{https://doi.org/10.1016/j.jvcir.2018.02.013}{doi.org/10.1016/j.jvcir.2018.02.013}}}
\setlength{\headsep}{0.35 in}


\section{Introduction}
\label{sec:introduction}

From the MylifeBits~\cite{MyLifeBits2002MM} project in the early 2000 until today, the processing of video data remains as one of the most challenging tasks for life-logging. Tasks such as acquisition, storage, and the proper use of large amounts of recorded data are particularly hard for video processing. Over the last couple of decades, technological advances in integrated circuits technology made available at low cost, high performance and low-power sensors and high performance processors with large capacity memory. As the energy storage capability improves, soon there will be cameras running all day. Indeed, mobile cameras such as GoPro\textsuperscript\textregistered, Looxcie, Google Glass\textsuperscript\texttrademark along with video sharing websites are boosting the popularity of the egocentric video in the last couple of years. More video data is being generated than ever before; daily an increasing number of users are filming themselves creating and sharing long video streams containing daily activities, such as walking, driving, cooking, sport activities, and working tasks.

Although egocentric videos have been widely produced and shared, they are hardly watched in their entirety because they are usually long and monotonous. Moreover, they contain jerky scene transitions causing visual unpleasantness and making it difficult to extract information from them~\cite{Poleg2015}. A simple approach for reducing the length of a video stream is by na\"ively sampling the video at every $n$-$th$ frame. However, this strategy severely impacts watching because it tends to amplify the jerkiness of natural body movements and to induce abrupt scenes transitions. These are just examples that make browsing and watching long videos a tough problem to solve.

Despite remarkable advances in processing techniques tasks such as video summarization, very few studies have addressed the problem of creating a pleasant experience to the watchers of egocentric videos. Recently, relevant efforts have been made to make these videos watchable~\cite{Kopf2014, Karpenko2014, Poleg2015, Joshi2015, Halperin2017}. Virtually all the proposed methodologies thus far were inspired by a photographic technique called \textit{Hyperlapse}. This technique aims at producing smooth videos from pictures taken of a scene with a selected and fixed point between shots. The works borrow from \textit{Hyperlapse} the idea of selecting only a subset of aligned frames in order to maximize the smoothness of the final video. 

The major drawback of the Hyperlapse inspired approaches is the assignment of a relevance score to each frame. They typically select a frame considering only the maximization of smoothness of the final video, thus neglecting the semantic content of each frame. In fact, as far as semantics are concerned, some parts of the videos may be more relevant than others for the user. For instance, in a video of a wedding, some specific moments are more charming, such as the bride's entrance, the family members close-ups, and the exchanging of vows and wedding rings. Hence, due to the skipping of stationary frames, characteristic of \textit{Hyperlapse} algorithms, relevant parts may be completely obliterated in the fast-forwarded version.

In this work, we propose a fully automatic multi-importance semantic fast-forward technique for egocentric videos designed to tackle with the challenging production of smooth fast-forward video without meaningful semantic loss. Our approach is composed of an adaptive frame selection and stabilization strategy. Our goal is to create a pleasant experience for the watchers preserving the continuity of the video and propitiating emphasis to relevant parts. 

Differently from our previous approaches~\cite{Ramos2016, Silva2016}, where we treated the semantic information as a binary problem, we manually fine-tuned the hyper-parameters and used an \textit{ad hoc} definition of semantic based on an existing classifier, in this work, we address these weaknesses by using:

\renewcommand{\theenumi}{\roman{enumi}}
\begin{enumerate}
	\item a new temporal segmentation approach, where the semantic information is organized in levels, turning the solution into a Multi-Importance approach;
	\item a fully automatic parameter setting that defines the speed-up and the weights of the graph via Particle Swarm Optimization;
	\item CoolNet: a Convolutional Neural Network trained to classify the semantic based on the preferences of the user. Therefore, predefining a specific semantic is not required;
	\item an in-depth analysis of the usage of different motion estimation techniques.
	
\end{enumerate}

\section{Related work}
\label{sec:related_work}

Video processing has been extensively studied in the past few years, mainly the video summarization problem. However, it is worth noting that video summarization and hyperlapse have important differences \cite{Molino2016}. On the one hand, Hyperlapse methods, as the one proposed by this work, are focused on creating a smooth fast-forward version of the input video, \ie, the output video is sped up entirely and unless they are too similar, no clips of the video are removed. On the other hand, video summarization methods are focused on creating compact visual summaries capable of presenting the most discriminative parts of the video as well as the most informative ones. 

\paragraph*{Video summarization}  
The goal of summarization techniques~\cite{Zhang2016, Song2016, Marvaniya2016} is to generate a shorter version of the video keeping the essential information by either creating  a static storyboard or still-image abstract, where some selected frames resume the relevant video content~\cite{Lee2012}, or a dynamic video skimming or moving-image abstract, where selected clips from the original stream are collated to compose the output video~\cite{Gong2000, Ngo2003}. 
\adding{Molino~\etal~\cite{Molino2016} perform an extensive study about the summarization of egocentric videos evidencing the importance of the area in an age of rising life-logging. One main issue is identifying the relevant information on the video, which could be subject to the recorder or to the viewer. }

As far as egocentric videos are concerned, only a few works have been developed recently~\cite{Lee2012, Lu2013, Lin2015, Xiong2015, Yang2016, Varini2017}. Lee~\etal~\cite{Lee2012} split the input video into temporal events based on color distribution and find relevant regions of a frame to compose a visual storyboard with the most important people and objects. Lu and Grauman~\cite{Lu2013} create a \textit{story-driven} summary by segmenting the video into sub-shots and detecting the key component of each sub-shot. Lin~\etal~\cite{Lin2015} propose a context-based highlight detection algorithm based on structured SVM to generate video highlights. 
\adding{Varini~\etal~\cite{Varini2017} proposed a method to customize the summarization regarding users preferences and GPS location by performing on-fly data gathering from online photos services and classifier training}. In spite of the fact that these techniques attain some sort of summarization of the relevant parts of egocentric videos, they produce, at best, only temporally discontinuous video sub-shots~\cite{Poleg2015}, since some otherwise relevant parts of the input video are completely left out.

\paragraph*{Hyperlapse} Hyperlapse strategies can be divided into two categories: 3D model approach, where methods aim at firstly creating the whole environment structure, and then finding the optimal path through the scene to create videos with smooth transitions, and; 2D only approach, which comprises methods focused on finding an optimal frame selection based on some smoothness criterion. 

A representative member of the 3D model category is the work by Kopf~\etal~\cite{Kopf2014}. Their method consists of three stages: scene reconstruction via structure-from-motion and per-frame proxy geometries; path planning by optimizing a 6D virtual camera path, and; image-based rendering via projection, stitching and blending of selected input frames. Despite the impressive results reported by this technique, it requires substantial scene overlap among frames and high computational cost. Moreover, if the scene parallax is small, it might generate numerous artifacts. 

The 2D only based methods~\cite{Karpenko2014, Poleg2015, Joshi2015, Ramos2016, Halperin2017} avoid this 3D reconstruction by sampling the frame of the input video optimally and decreasing the processing. 
The Hyperlapse from Instagram~\cite{Karpenko2014} combines gyroscope samples and frames into a stabilizer to obtain the camera orientations which are fed into a video filtering pipeline to obtain steady frames. Poleg~\etal~\cite{Poleg2015} create a graph from the input video taking the frames as nodes and edges values as a linear combination of shakiness, speed of motion and appearance between pairs of frames. Their final video is composed of those frames related to nodes of a shortest path. Recently, Halperin~\etal~\cite{Halperin2017} extended the approach by Poleg~\etal's  with an expansion of the field of view of the output video. They use the mosaicking technique on the input frames of one or more egocentric videos.
The Microsoft Hyperlapse algorithm~\cite{Joshi2015} optimally selects the set of frames via dynamic-time-warping which present the smoothest transitions with homography transformations. 

Even though the aforementioned techniques have succeed in creating smooth fast-forward output videos, they do not take into account specific user interests on watching such videos. Some segments of the videos may have scenes with different relevance for the recorder.

Recently, Yao~\etal \cite{Yao2016} proposed learning the relationship between paired highlights and non-highlights segments to create a summary of the video.  
Although the work is focused on video summarization, it has a twofold purpose. The method returns a composition of skims and a video timelapse. The authors use the timelapse to estimate the rate to play the highlight segments in slow motion. The remaining segments are played in a fast-forwarded manner to achieve the final length, shorter than the original. It is noteworthy that the authors assume the number of highlight segments smaller than the number of non-highlight segments. \cutoff{This assumption does not hold for those cases such as the `Biking 50p', `Driving 50p' and `Walking 75p' videos in the semantic dataset proposed by Silva~\etal~\cite{Silva2016}.}
When compared to Yao~\etal's work, our methodology is a lighter and presents a more modular approach since we use the confidence given by classifiers and a threshold to identify the relevance and the boundary of the segment. Our approach is based on an adaptive frame selection, focusing on choosing frames that lead to a more stable video, while they use the uniform sampling approach.  Also, our segmentation strategy is capable of handling different configurations for the highlights lengths.

\adding{Higuchi~\etal~\cite{Higuchi2017} proposed a fast-forwarding video interface for users to browse the important events on first-person videos quickly, confirming the relevance of emphasizing semantic information. The user set egocentric cues (hand, ego-motion or face), which are used to play at normal speed segments of the video containing theses cues. The remainder of the video is played faster, using an uniform sampling, in a speed-up rate selected by the user.}

In our previous work~\cite{Ramos2016}, we used a semantic threshold to classify the input frames as semantic or non-semantic. Then, we split the video into segments of those types and calculate different speed-up rates for each type of segment such that the semantic segments are emphasized by a lower speed-up. We later extended this approach by improving the slicing strategy with a new thresholding method and introducing a new egocentric stabilization process~\cite{Silva2016}. We also proposed a semantically labeled dataset and defined an instability metric for egocentric videos.

Unlike our previous works, in this work, we use a multi-importance approach in our splitting strategy to segment the video temporally. In other words, rather than labeling the segments as semantic or non-semantic, we define multiple levels for the semantic segments. Thus, the segments can be emphasized according to their importance. We also seek the automation of some steps. We use the Particle Swarm Optimization (PSO)~\cite{Kennedy1995} algorithm to better select the weights for the speed-ups definition and the transition costs of the frames. Furthermore, we remove the need for an existing classifier to determine the semantic score.

\section{Methodology}
\label{sec:methodology}

Our methodology is composed of two main steps: (i) identifying and selecting frames adaptively, and; (ii) stabilizing the fast-forward video.

\subsection{Semantic egocentric fast-forwarding}
\label{subsec:methodology_frame_sampling}

\begin{figure}[t!]
	\centering
	\includegraphics[width=\textwidth]{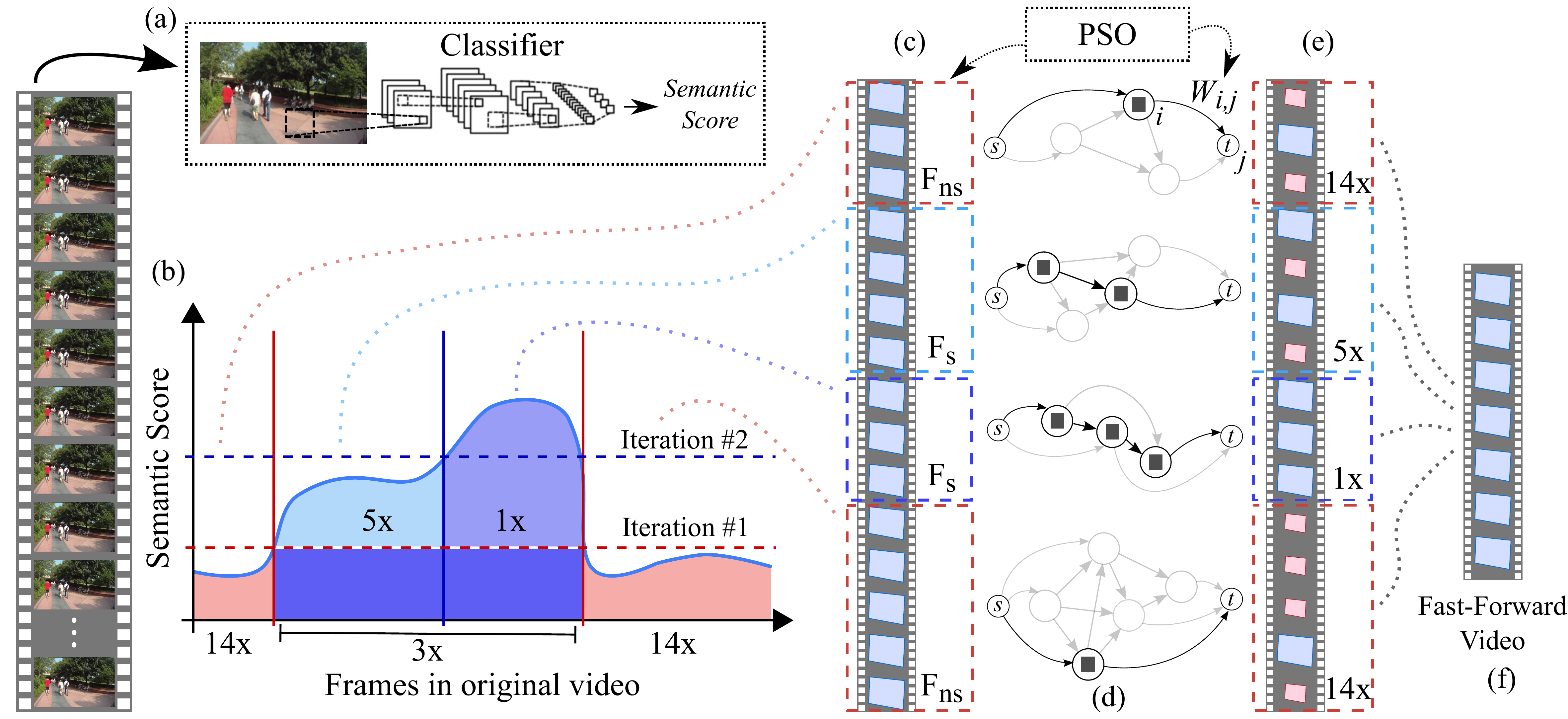}
	\caption{Frame Sampling: We extract the semantic information to create a score \adding{for each frame} (a) creating a video profile used to split the video into non-semantic and semantic segments, and then refining the semantic segments, iteratively (b). We calculate speed-up rates for each type of segment\adding{, emphasizing the most relevant by a lower rate} (c). We create a graph for each segment and calculate the shortest path (d). Finally, the selected frames (e) compose the output video (f). The PSO algorithm is used to optimize steps (c) and (d).}
	\label{fig:methodology}
\end{figure}

The adaptive frame sampling of our methodology is composed of five steps. We first extract the semantic information (\eg, people, car plates, charming environments) from each frame of the input video. These data define a semantic profile of the video which we use to split the stream into relevant and non-relevant segments. For each type of segment, we calculate different speed-up rates, assigning lower rates to the relevant segments. We build a graph for each segment of the video where the frames are the nodes and the temporal relation between two frames is defined by an edge. The edge weight is related to the cost of keeping the frames sequentially in the fast-forward video. We then run a shortest path algorithm to find the selection with the smaller transition costs and generate the final video. 

\paragraph*{Semantic extraction} The semantic information is encoded by the score function $S: \mathbb{R} \rightarrow \mathbb{R}$, which is composed of three components: (i) the confidence of the extracted information, which is given by the classifier (\eg, a face detector, a pedestrian detector); (ii) the position with respect to the center of the analyzed region -- as the input is an egocentric video, the central area of the frame should have a higher relevance to the viewer, and; (iii) the size of the region since, in general, larger areas means closer objects, therefore it represents a higher probability of interaction. 

\cutoff{Let $k$ be the $k$-th Region of Interest (ROI) returned by a classifier for the frame $f_x$ of dimensions $W \times H$} \adding{In this work, we assume that the classifier is able to return both the classification confidence and the Region of Interest (ROI). Thus, let $k$ be the $k$-th ROI extracted by a classifier for the frame $f_x$ of dimensions ${W \times H}$}. \cutoff{To quantify the centrality of the object, we use a Gaussian mask centered at the frame $f_x$ with standard deviation $\sigma=\min(W/2, H/2)$} \adding{To quantify the centrality of the object, we use a Gaussian mask, $G_{\sigma}(k)$, centered at the frame $f_x$ with standard deviation $\sigma=\min(W/2, H/2)$.}
Higher values are assigned to objects closer to the central point of the frame. 
The semantic score is given by:
\begin{equation}
\label{eq:semantic_score}
S_{x} = \sum_{k \in f_{x}} c_{k} \cdot a_{k} \cdot G_{\sigma}(k),
\end{equation}
\noindent where $a_{k}$ is the normalized area size in pixels of the $k$-th ROI and $c_{k}$ is the normalized confidence returned by the classifier for the ROI $k$. It assigns a relevance proportional to the reliability of the semantic information.

We also classify the semantic contents of a video using a Convolutional Neural Network (CNN) based on the preference of the user. In this work, we propose the \textit{CoolNet}, a network model used to rate the frame ``coolness'' based on web video statistics. To achieve our goal, we train a CNN analyzing the frame in its entirety, similarly to the scene recognition problem. Therefore, we propose to use the VGG16 model trained on MIT Places205 dataset~\cite{Zhou2014} fine-tuned in our domain. 
\cutoff{When using the CNN, the semantic score $ S_{x} $ of a frame $ f_{x} $ is given by the network output when analyzing the frame} \adding{When using the CNN, the semantic score $ S_{x} $ of a frame $ f_{x} $ is given by the network output alternatively to the Equation~\ref{eq:semantic_score}}.

\paragraph*{Temporal segmentation} The sequence of semantic scores computed for each frame defines the semantic profile of the video as illustrated in Figure~\ref{fig:methodology}-b. We split the video to create temporal segments by thresholding the semantic profile. We create a histogram with the semantic scores and use the Otsu thresholding method~\cite{Otsu1979} to define the semantic threshold. Thus, every frame above this threshold is labeled as a semantic frame. Consecutive frames labeled as semantic compose the semantic segments and the remaining frames compose the non-semantic segments.

\paragraph*{Speed-up rate estimation} Differently from the works of Ramos~\etal~\cite{Ramos2016} and Silva~\etal~\cite{Silva2016} that estimate two speed-up factors: one for semantic segments and the other for non-semantic segments, in this work we present a multi-importance approach. Our approach is capable of working with \cutoff{different} \adding{many} levels of semantic content by estimating \cutoff{a different speed-up} \adding{unique speed-up rates} for each semantic segment according to the scores of its frames.

Initially, the temporal segmentation step is executed once to compute the speed-up that will be used in non-semantic parts, and then we iteratively segment the semantic parts for refining the semantic speed-up, as illustrated in Figure~\ref{fig:methodology}-b. In each iteration, we decrease the speed-up rate for emphasizing the segments with higher semantic content. Estimating the speed-ups is a non trivial task, since the total length of the semantic segments may vary a great deal. Additionally, the final speed-up for the whole video should be closer to the desired speed-up.

Let $F_{d}$ be the speed-up rate chosen by the user, $L_{s}$ the total number of frames in all semantic segments and $L_{ns}$ the number of frames in non-semantic segments. We compute the semantic speed-up $F_{s}$ and the non-semantic speed-up $F_{ns}$ by minimizing the energy function:
\begin{equation}
\label{eq:desired_speedup}
D(F_{ns}, F_{s}) = \left|\frac{L_{s}+L_{ns}}{F_{d}} - \left(\frac{L_{s}}{F_{s}} + \frac{L_{ns}}{F_{ns}} \right) \right|.
\end{equation}
We include the speed-up $F_{s}$ and the difference between $F_{ns}$ and $F_{s}$ as regularization terms for helping finding a minimum of Equation~\ref{eq:desired_speedup}. Because for every $F_{s}$ value there is a $F_{ns}$ leading the result to $0$, we have more the one solution which can minimize Equation~\ref{eq:desired_speedup}.
Also, to create a finite and discrete search space, we use \adding{a set of constraints $\mathcal{R}$ composed of:} \cutoff{three restrictions: (i)} \adding{($r_1$)} $F_{s} \leqslant F_{d}$, because we want  emphasis in the semantic parts; \cutoff{(ii)} \adding{($r_2$)} $F_{ns} \geqslant F_{d}$, since we want to achieve desired speed-up in the fast-forward video; and \cutoff{(iii)} \adding{($r_3$)} $F_{s} \geqslant p_{s} F_{d}$, where $p_{s} = L_{s}/(L_{s} + L_{ns})$, to avoid an excessive number of frames. Thus, the optimization problem is given by:
\begin{equation}
\label{eq:argmin}
\begin{split}
&\argmin{F_{s},\,F_{ns}} D\left(F_{ns}, F_{s}\right) + \lambda_{1} |F_{ns}-F_{s}| + \lambda_{2} |F_{s}|\\
&\text{subject\,to}\, r_i \in \mathcal{R} \\
\end{split}
\end{equation}
\noindent where $\lambda_{1}$ and $\lambda_{2}$ are the regularization parameters used to control the importance of keeping the speed-up rates close or taking the smaller $F_{s}$, respectively.

\adding{We applied an iterative multi-relevance approach to refine the values of $ L_{s} $, $ L_{ns} $, and $ F_{d} $ in Equation~\ref{eq:desired_speedup}. Firstly, we compute $ F_{s} $ and $ F_{ns} $. For each new iteration, we remove the non-semantic segments and create a new semantic profile for the video. The $ F_{s} $ value is used as the newly required speed-up, preserving the overall $ F_{d} $. The process stops when the new semantic threshold is lower than the last threshold times $t$. This stop condition avoids creating a large number of segments.}

\paragraph*{Graph building} We build graphs, similar to the work of Halperin~\etal~\cite{Halperin2017}, one for each segment with each node connected with $\tau_{max}$ subsequent frames (Figure~\ref{fig:methodology}-d). The weight $W_{i,j}$ of the edge that connects the $i$-th to $j$-th node\adding{s} is given by the linear combination of the terms related to the frames transition instability $I_{i,j}$, appearance $A_{i,j}$, velocity \cutoff{$V{i,j}$} \adding{$V_{i,j}$} and semantic $S_{i,j}$ multiplied by a weighting factor, as shown in Equation~\ref{eq:graph_formulation}:
\begin{equation}
\label{eq:graph_formulation}
W_{i,j} = (\lambda_{I} \cdot I_{i,j} + \lambda_{V} \cdot V_{i,j} + \lambda_{A} \cdot A_{i,j} + \lambda_{S} \cdot S_{i,j}) \cdot \left\lceil{\frac{(j-i)}{F}}\right\rceil.
\end{equation} 
\noindent The weighting factor enhances transitions between frames with lower distance and $F$ is the speed-up rate applied in the graph which the edge belongs. The $\lambda$ coefficients are regularization factors for the cost terms. 

\adding{Smooth transitions are rewarded by the instability cost term, $I_{i,j}$, calculated by the average distance of the Focus of Expansion (FOE) to the center of the image. The velocity cost term, $V_{i,j}$, indicates the speed sensation and it is given by the difference between the average magnitude of the optical flows (OF) of the whole video and the OF along the consecutive frames from $i$ to $j$. The similarity appearance between the frames $i$ and $j$, represented by the term $A_{i,j}$, is calculated using the Earth Mover's Distance of the color histogram of both frames.}

We penalize the transitions that are not composed of frames with relevant semantic information trough the Semantic Cost Term, which is computed as $S_{i,j} = \frac{1}{S_{i} + S_{j} + \epsilon}$, where $S_{i}$ and $S_{j}$ are the semantic scores for the frames $ i $ and $ j $, respectively, and $ \epsilon $ avoids dividing by zero when both scores are null.

\paragraph*{Automatic parameter setting} As it can be seen, in Equations~\ref{eq:argmin}~and~\ref{eq:graph_formulation} there are a total of six parameters highly related to the input video. Since their values are continuous, the search space is very large. Their configuration demands much user knowledge and effort, moreover there is a high probability that the user will stop before finding the right parameters. Setting up universal parameters as described in Poleg~\etal~\cite{Poleg2015} may not be the best approach, which can be confirmed by analyzing their results (EgoSampling) in Section~\ref{sec:experiments_evaluation_metrics}. We propose a parameter setting via PSO to automate the selection of the parameter values.

The PSO algorithm is an iterative method that groups particles arranging them randomly in the search space. At every iteration, the particles positions (parameters values) are updated to follow the local and global best particles. The solution is given by a fitness equation defined according to the problem. In our case, we define the following fitness equation:
\begin{equation}
\fitness{\lambda_{1},\lambda_{2}} = c \cdot \left|\widehat{F_{s}} - \frac{F_{d} + p_{s} \cdot F_{d}}{2}\right| + |\widehat{F_{d}}-F_{d}| + p_{ns} \cdot |\widehat{F_{s}} - \widehat{F_{ns}}|,
\label{eq:fitness_argmin}
\end{equation}
\noindent which estimates $\lambda_{1}$ and $\lambda_{2}$ of Equation~\ref{eq:argmin}. The $ \widehat{F_{s}} $ and $ \widehat{F_{ns}} $ are the best values of $ F_{s} $ and $ F_{ns} $ in the finite and discrete search space when replacing $\lambda_{1}$ and $\lambda_{2}$ with the particle position. The value $p_{s} = L_{s}/(L_{s} + L_{ns})$ is the semantic percentage of the video,  $p_{ns} = L_{ns}/(L_{s} + L_{ns})$ is non-semantic percentage, $ c = 2 $ is a constant value to control the importance of selecting a lower semantic speed-up and $ \widehat{F_{d}} = (L_{s} + L_{ns})/(L_{s}/\widehat{F_{s}} + L_{ns}/\widehat{F_{ns}}) $ is the speed-up achieved with the selected speed-ups.

For the remaining parameters $ \lambda_{I} $, $ \lambda_{V} $, $ \lambda_{A} $ and $ \lambda_{S} $ in the Equation~\ref{eq:graph_formulation}, we use the fitness equation:
\begin{equation}
\fitness{\lambda_{I},\lambda_{V},\lambda_{A},\lambda_{S}} = \frac{J}{Max_J} + \left|\frac{\widehat{L} - E_{L}}{E_{L}}\right| + \frac{\widehat{S}^{*} - Semantics}{\widehat{S}^{*}},
\label{eq:fitness_graph_formulation}
\end{equation}
\noindent where $J$ is the jitter of the generated fast-forward video, $ Max_J $ is the maximum possible jitter for the video, $ E_{L} $ is the expected number of frames, $ \widehat{L} = L / \widehat{F_{d}} $ is the final video length, $ L $ is the original video length, and $ \widehat{S}^{*} $ is the maximum value for the semantic score of the fast-forward video. 

The $Semantics$ value represents the semantic content of the generated fast-forward video. It is the sum of the semantic score computed by the Equation~\ref{eq:semantic_score} using all frames. We compute the jitter as the magnitude of the mean deviation of the FOE locations along the selected frames and the maximum possible jitter is the jitter of a hypothetical video where for every frame the FOE is as far as possible from the previous.

\paragraph*{Video composition} The last step of our methodology is \cutoff{connecting the graphs with the $\tau_{b}$ border frames and computing} \adding{adding source and sink nodes for each graph, and connecting them to $\tau_{b}$ border frames with zero-weighted edges. We compute the} shortest path (Figure~\ref{fig:methodology}-d) using the Dijkstra algorithm. All frames related to the nodes within the shortest path will compose the final video as depicted in Figure~\ref{fig:methodology}-f.

\subsection{Egocentric video stabilization}
\label{subsec:methodology_video_stabilization}

As noted by Kopf~\etal~\cite{Kopf2014}, traditional video stabilization algorithms do not perform well on \cutoff{egocentric} \adding{first-person} videos. This can be assigned to the difficulty in tracking the motion between successive frames, which is increased in the fast-forward version. Therefore, we stabilize our semantic fast-forward version of the original video in a way similar to \adding{the work of} Silva~\etal~\cite{Silva2016}.

\begin{figure}[t!]
	\centering
	\includegraphics[width=1.0\textwidth]{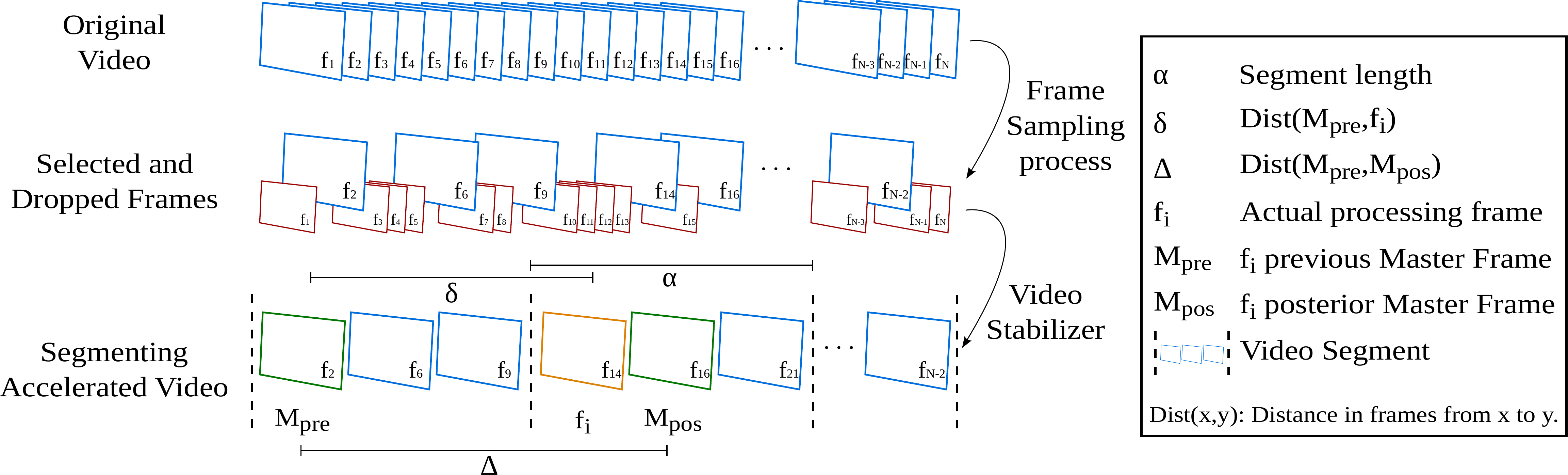}
	\caption{Video Stabilization: The top row depicts the original video in a frame sequence. The middle row shows the selected and dropped frames (larger blue and smaller red frames, respectively) in the sampling process. The last row \cutoff{gives} \adding{presents} an example of the fast-forward video segmentation, \cutoff{with the} master frames\adding{,} and \cutoff{the} terms $\alpha$, $\Delta$ and $\delta$.}
	\label{fig:stabilization_method}
\end{figure}

Using the frames from the sampling step of Section~\ref{subsec:methodology_frame_sampling}, we first split the video into segments of size $\alpha$ and select one master frame for each segment (Figure~\ref{fig:stabilization_method}). A master frame $M_{k}$ \cutoff{of the $k$-th segment} is the frame $f$ that maximizes the
number of \textit{inliers} obtained with RANSAC when computing the homography transformation from the image \cutoff{$x$ to $y$} \adding{$f$ to all images into the $k$-th segment}.

\begin{algorithm}[t!]
	\caption{Egocentric Fast-Forward Video Stabilizer}
	\label{alg:VideoStabilizer}
	\scriptsize{
		\begin{algorithmic}[1]
			\Require Set of frames $\mathcal{V}$ in the fast-forward video; Set $\mathcal{D}$ of dropped frames in the sampling process; The $crop\_area$ and $drop\_area$.
			\Ensure The set of stabilized frames $\mathcal{S}$.		
			\Function{VideoStabilizer}{$\mathcal{V}$, $\mathcal{D}$}
			\State $\mathcal{S} \gets \{\}$
			\ForAll{$f_{i}$ $\in$ $\mathcal{V}$}
			\State $ w \gets \left( \delta \cdot \left( 2 \cdot \alpha \right) / \Delta \right)  $ 
			\State $\widehat{f}_{i} \gets H_{f_{i},M_{pre}}^{1-w} \cdot H_{f_{i},M_{pos}}^{w} \cdot f_{i} $ \label{alg_line:new_fi}
			\While{$\widehat{f}_{i} \cap \textit{crop\_area} < \textit{crop\_area}$}
			\label{alg_line:reconstruction_process_start}
			\If {$\widehat{f}_{i} \cap \textit{drop\_area} = \textit{drop\_area}~\textbf{and}~\textit{ExistUnusedFrames}(\mathcal{D}) $}
			\State $\widehat{f}_{i} \gets \textit{Stiching}\left(\widehat{f}_{i},GetUnusedFrame(\mathcal{D})\right)$ \label{alg_line:stiching}
			\Else
			\State ${f}_{d} \gets \textit{SelectNewFrame}\left( \mathcal{D} , {f}_{i} \right)$ \label{alg_line:select_new_frame}						
			\State $ w \gets \left( \delta \cdot \left( 2 \cdot \alpha \right) / \Delta \right)  $ \Comment{\small{Recalculate distances using ${f}_{d}$ as ${f}_{i}$.}}
			\State $\widehat{f}_{i} \gets H_{f_{i},M_{pre}}^{1-w} \cdot H_{f_{i},M_{pos}}^{w} \cdot {f}_{d} $
			\EndIf
			\label{alg_line:reconstruction_process_end}
			\EndWhile
			\State $ \mathcal{S} \gets \mathcal{S} + \{\widehat{f}_{i} \cap crop\_area \} $
			\EndFor
			\EndFunction
		\end{algorithmic}
	}
\end{algorithm}

\cutoff{In the second step we smooth out the frame transitions. As shown in line~\ref{alg_line:new_fi} of the Algorithm~\ref{alg:VideoStabilizer}, for} 
\adding{In the second step, we smooth out the frame transitions following the steps in Algorithm~\ref{alg:VideoStabilizer}. For} each frame $f_{i}$ of the fast-forward video, we compute a frame $\widehat{f_{i}}$ of the stabilized video using $\widehat{f_{i}} = H_{f_{i},M_{pre}}^{1-w} \cdot H_{f_{i},M_{pos}}^{w} \cdot f_{i}$ \adding{(Algorithm~\ref{alg:VideoStabilizer}~-~line~5)}. The \cutoff{matrices} $H_{f_{i},M_{pre}}$ and $H_{f_{i},M_{pos}}$ \adding{are homography matrices which take} \cutoff{transform} the frame $f_{i}$ to the \adding{image plane of the} previous master frame $M_{pre}$ and to the posterior master frame $M_{pos}$, respectively. The $\delta$ value is number of frames from $f_{i}$ to $M_{pre}$, and $\Delta$ is the number of frames between $M_{pre}$ and $M_{pos}$. 
Like Hsu~\etal~\cite{Hsu2012}, we weight the homography transformations according to the distance to the master frames. 

Black areas may be created after applying the homographies because of abrupt motions of the camera and the large elapsed time between consecutive frames in the fast-forward videos. Thus, we define two areas centered in the frame to decide when a frame should be reconstructed: the drop area, which is equal to $dp\%$ size of the frame and the crop area equals to $cp\%$ size of the frame ($cp > dp$). The reconstruction asserts that every $\widehat{f}_{i}$ frame covers the crop area.

There are two conditions for reconstructing a frame: (i) $\widehat{f}_{i}$ does not create black regions in the central area; and (ii) there are unused frames in the dropped set $\mathcal{D}$ for stitching. If both these conditions hold, we perform the stitching using $\widehat{f}_{i}$ and a new frame from $\mathcal{D}$. If at least one of the conditions is false, we discard $\widehat{f}_{i}$ and select a new frame ${f}_{d}$ from $\mathcal{D}$ and recalculate the distances and the homography matrices. Once the crop area is covered, the intersection between this area and the frame $\widehat{f}_{i}$ compose the $i$-$th$ frame in the stabilized video.

If the $\widehat{f}_{i}$ does not yield a good transition in the final video, we select a new frame $f_{d}$  belonging to the interval $[f_{i-1}, f_{i+1}]$ in the set of dropped frames $\mathcal{D}$ (Algorithm~\ref{alg:VideoStabilizer}~-~line~\ref{alg_line:select_new_frame}) and that maximizes the equation:
\begin{equation}
\label{eq:select_new_frame}
\underset{f_{d}}{\arg\max}~(~G_{\sigma}(p) \cdot (~R(f_{d},{f}_{i-1}) + R(f_{d},{f}_{i+1})~) \cdot (\eta + S(f_{d}))~),
\end{equation}
\noindent where $G_{\sigma}(x)$ is a Gaussian function with mean ${\mu = 1}$ and standard deviation $\sigma$ in the position $x$, $p$ is the percentage of the crop area covered by $\widehat{f}_{d}$, $R (.)$ is the number of \textit{inliers} obtained with RANSAC and $S(.)$ is the semantic score given by Equation~\ref{eq:semantic_score}. The term $\eta$ is used to avoid multiplying by zero.

\section{Experiments}
\label{sec:experiments}

In this section, we present an experimental evaluation, which includes describing the datasets used, the parameters configuration, weights setting, methods and metrics chosen for quantitative comparison, and the result discussion. \cutoff{We also discussion what makes some information a good semantic for general propose and how to learn and extract it} \adding{Furthermore, we discuss the meaning of the semantics for general purposes and how to learn it}. 

\paragraph{Dataset}
\label{subsec:experiments_semantic_egocentric_dataset}

We use two datasets to evaluate our methodology. The first one is a composition of standard sequences used in validating \adding{previous} egocentric methods: Bike 1, Bike 2, Bike 3, Walking 1 and Walking 2 from~\cite{Kopf2014}; Running, Driving and Walking 3 from~\cite{Poleg2015} and; Walking 4~\cite{Poleg2014}. Hereinafter referred to as {\it Unlabeled Dataset} because there is no annotation of its semantic content.

The second dataset, referred to as {\it Semantic Dataset}\footnote{Publicly available at:  \href{https://www.verlab.dcc.ufmg.br/semantic-hyperlapse/epic2016-dataset}{www.verlab.dcc.ufmg.br/semantic-hyperlapse/epic2016-dataset/}}, was recently presented in our previous work~\cite{Silva2016}. It is composed of $11$ sequences of $3$ different activities: Biking; Driving and Walking. 
\cutoff{All the sequences are labeled as being: 0p, for videos with approximately no semantic information (Biking 0p, Driving 0p and Walking 0p); 25p, for the videos containing relevant semantic information in $25\%$ of its frames (Biking 25p, Driving 25p and Walking 25p); 50p, for the ones with around a half of their frames has some semantic content (Biking 50p, Biking 50p2, Driving 50p and Walking 50p) and; 75p for the videos with $75\%$ of their frames containing relevant semantic (Walking 75p).}
\adding{The sequences are labeled with the suffix 0p, 25p, 50p, or 75p, for videos with approximately no semantic information, cointaining semantinc content in approximately $25\%$, $50\%$, or $75\%$ of frames, respectively. As mentioned in the Section~\ref{sec:related_work}, the work of Yao~\etal can not handle videos in which the semantic portions are longer than the non-semantic ones. Videos named with the suffix 50p and 75p are examples of cases which the Yao~\etal method fails.}
\adding{The semantic used were faces for Walking videos and pedestrians for Driving and Biking ones. A frame is labeled as semantic if its score is higher than the video semantic threshold, as described in Section~\ref{subsec:methodology_frame_sampling}.}

\paragraph{\cutoff{Parameters Setup} \adding{Implementation details}}
\adding{Since we have added strong space restrictions to the optimization function represented in Equation~\ref{eq:argmin}, we solve it by exhaustive search. The same approach is used to solve the maximization problem expressed by Equation~\ref{eq:select_new_frame}, once the numbers of frames is small into the segment $[f_{i-1},f_{i+1}]$.}
We empirically set the parameters of our methodology, following a careful procedure to \cutoff{insure} \adding{ensure}, as much as possible, the best overall results.  For the experiments on evaluating the semantic content, we used the NPD Face Detector~\cite{Liao2016} and a pedestrian detector~\cite{Nam2014} as the semantic extractors.

\adding{For the following parameters, we test in a subset of videos and keep the same values for all videos of both datasets.} Thresholds of $c_k = 60$ and $c_k = 100$ are used as minimum value for the confidence of an accepted face detection and for pedestrian detections, respectively. \adding{The values were set to prevent false positive detections. }
In the temporal segmentation (Section~\ref{subsec:methodology_frame_sampling}), we filter the semantic profile using a Gaussian function with \cutoff{$\sigma = 5 \cdot fps$} \adding{$\sigma = s_d / 2 \cdot fps$}, where \adding{$s_d$ and} $fps$ stand for required speed-up and frames per second, respectively. We only consider ranges greater than \cutoff{$5$ seconds} \adding{$1$ second in the accelerated video}, since short ranges would result in \cutoff{short} a flash in the final video. Likewise, we connected every range distant by $5$ seconds or less. 
For the graph building step, we set the values \adding{as recommended by the authors of the work Poleg~\etal}, the border frames $\tau_{b} = 30$ and the maximum allowed skip $\tau_{max} = 100$. In the semantic cost term equation we use $\epsilon=1$ \adding{to avoid division by zero}. 

In the video stabilization step, \adding{we perform a grid search for setting the values of $\alpha$, $\eta$, and $\sigma$. The values which lead to a better trade-off between the final video stability and the computation time are $\alpha = 4$, $\eta = 0.5$, and $\sigma = 10$} 
\cutoff{the size of the segments for selecting the master frames was defined as $\alpha=4$. The area of $da$ as $dp = 50\%$ of the frame and the area of $ca$ as $cp = 90\%$. 
	In the Equation~\ref{eq:select_new_frame}, we use  $\eta = 0.5$ and $\sigma = 10$.}

\subsection{Quantitative analysis}
\label{sec:experiments_evaluation_metrics}

The main goal of this work is to automatically create visually pleasant fast-forward videos and to emphasize the segments that are rich in semantic content. 
Therefore, we performed experiments analyzing both the amount of semantic content, the video smoothness, and the overall speed-up rate. We compare our methodology against the Stabilized Semantic Fast-Forward (SSFF)~\cite{Silva2016}, EgoSampling (ES)~\cite{Poleg2015} and Microsoft Hyperlapse (MSH)~\cite{Joshi2015}. 

\paragraph{Semantic evaluation}

\begin{figure}[t!]
	\centering
	\includegraphics[width=1.0\textwidth]{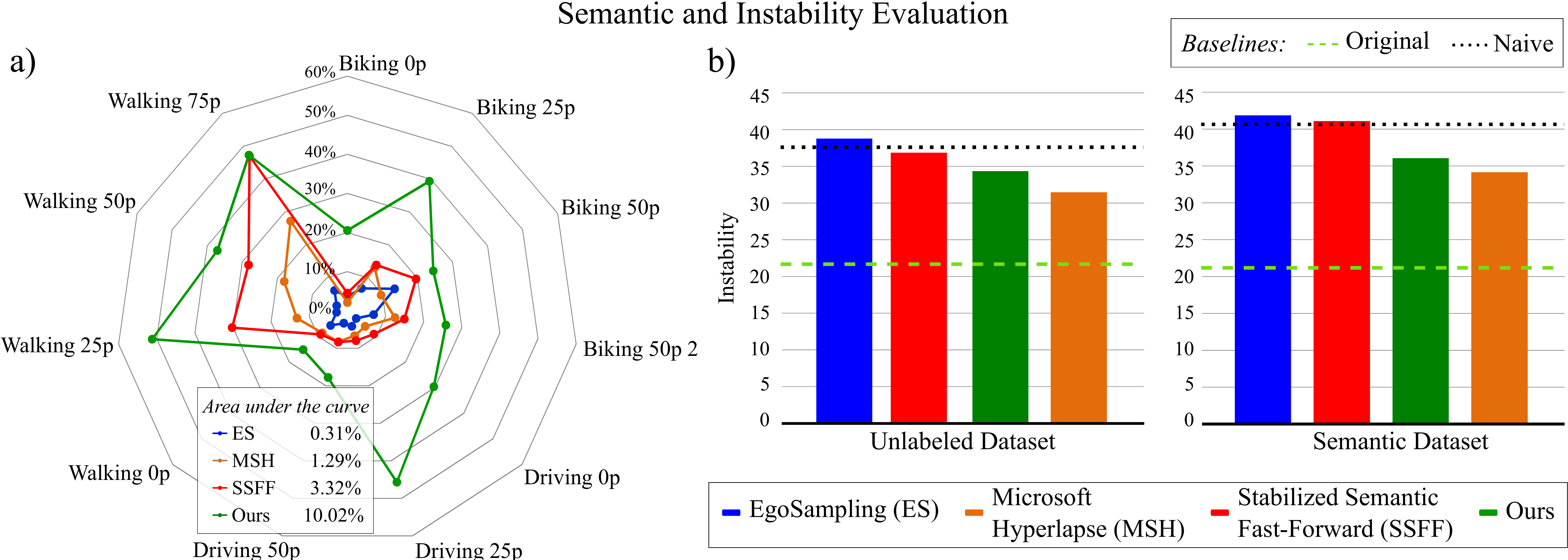}
	\caption{Semantic and Instability evaluation against the state-of-the-art methods.}
	\label{fig:Instability_and_Semantic_Evaluation}
\end{figure}

Figure~\ref{fig:Instability_and_Semantic_Evaluation}-a shows the fraction of the semantic content retained from the maximum value that can be present in a fast-forward video. We calculate this maximum by summing over the $ n $ top-ranked frames with relation to the semantic content, where $n$ is the ratio of the accelerated video length by the speed-up rate required.

It can be seen in Figure~\ref{fig:Instability_and_Semantic_Evaluation}-a that in most cases, our proposed method leads the performance. The new multi-importance approach is the main reason of this superiority, once it allows the creation of videos with even more emphasis in the semantic segments with higher semantic content. In the Driving 25p case, we manage to keep around $ 50\% $ of the possible semantic information, while the state-of-the-art semantic fast-forward method, SSFF, only takes $ 10\% $, what means $5$ times more semantic information in the final video. Biking 25p and Driving 0p are also cases with around $3$ times more semantic content compared to SSFF. For the Walking 75p sequence our multi-importance approach achieves similar value of SSFF. It creates one semantic clip probably because of the low variation along the whole semantic segment. 

Our methodology manages to keep over $3$ times more semantic content than the SSFF method, which is also a semantic fast-forward method. In comparison to the MSH, which is the best non-semantic fast-forward technique, the average semantic information kept is $8$ times higher.

\paragraph{Instability evaluation}
\label{sec:instability_metric}

One side effect of the semantic fast-forward is the increasing of the shakiness in the non-semantic segments. In general, the speed-up rate in these segments is higher than the desired, once semantic segments are emphasized by a low speed-up rate. Moreover, the higher the speed-up rate to a segment, more difficult is the stabilization, since consecutive frames may contain a small overlap.

Most of the fast-forward methodologies either used qualitative metrics, which involve human evaluation on the videos, or the epipole/FOE jitter metric in the final video as the quantitative instability evaluation. In our previous work~\cite{Silva2016}, we showed that this metric occasionally assigned better scores for shakier videos, considering the preference of the users. In this work, we evaluate the smoothness of the final produced video using the metric inspired by the qualitative comparison between videos made by Joshi~\etal~\cite{Joshi2015}, which uses side-by-side comparisons, calculating the mean and the standard deviation frames of consecutive images~\cite{Silva2016}. The metric is defined as follows:
\begin{equation}
\label{eq:instability_index}
I = M\left(\frac{1}{N} \cdot \sum_{i=1}^{N} \sqrt{\frac{\sum_{j \in B_i}~(f_{j} - \bar{f_{i}})^2}{(N_{B}-1)}}\right),
\end{equation}
\noindent where $N$ is the number of frames in the video, $B_i$ is the $i$-th buffer composed by $N_{B}$ temporal neighborhood frames, $f_j$ is the $j$-$th$ frame of the video, $\bar{f_i}$ is the average frame of the buffer $B_i$, $ M(\cdot) $ is a function that returns the mean value for the pixels of a given image and $I$ indicates the instability index of the video. A smoother video yields a smaller $I$ value.

Figure~\ref{fig:Instability_and_Semantic_Evaluation}-b depicts the average instability of the videos generated by the four techniques and two baseline methods. We consider the original and na\"ive sampling videos as baselines. The original videos are used as a baseline for the best smoothness, while the na\"ive sampling are used as a baseline for a poor result. 

According to the results in Figure~\ref{fig:Instability_and_Semantic_Evaluation}-b, our method is slightly below the best result. As expected, the MSH technique has the lowest instability value, since it aims at optimizing the stability of the fast-forward version. Our methodology, for its turn, aims at emphasizing the semantic segments. Therefore, it prefers dropping frames that result smoother transitions than removing those with higher semantics. Nevertheless, the videos generated by our methodology are preferable over the ES and SSFF ones. 

\paragraph{Speed-up evaluation}
\label{sec:other_evaluation_metrics}

In this experiment we verify whether the output videos lengths are close to the speed-up chosen by the user. We calculate the speed-up rate of the output videos of the techniques in both datasets. The speed-up rate is given by the ratio between the number of frames in the input video and the number of frames in the output video. 
The videos created by our methodology have an average absolute difference to the required speed-up of $0.25$ against $0.74$ of MSH, $3.00$ of SSFF and $10.97$ of EgoSampling. It also presents the lowest standard deviation of the absolute differences which is $0.78$ against $0.86$ of MSH, $2.46$ of SSFF, and $6.98$ of ES. 

\paragraph{Weights setup}

\begin{figure}[ht!]
	\centering
	\includegraphics[width=0.9\textwidth]{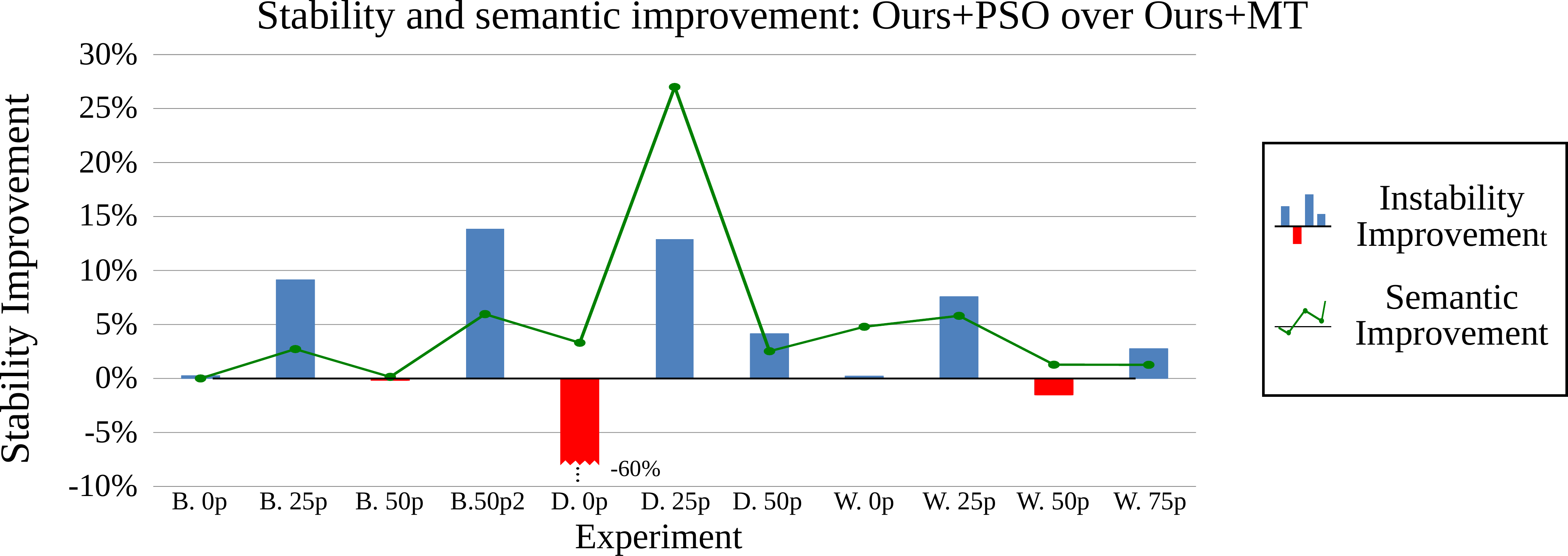}
	\caption{Stability and semantic improvement using PSO (Ours+PSO) over the best Manual Tuning (Ours+MT). A $100\%$ of improvement in instability means a video as stable as the original. The W., D., and B. stand for Walking, Driving and Biking experiments, respectively.}
	\label{fig:PSO_vs_HFT_Evaluation}
\end{figure}

To verify that the parameter setting contribute to the success of our approach, we also compared the results of our methodology using Particle Swarm Optimization (Ours+PSO) with using the best Manual Tunning (Ours+MT). Figure~\ref{fig:PSO_vs_HFT_Evaluation} depicts the improvement of the instability and semantic content of Ours+PSO over Ours+MT for the output videos of the Semantic Dataset. Together, these results show that the use of PSO for the parameter setting is a positive contribution to our methodology, either in the semantic content or video instability. We credit these results for the power of convergence of the PSO algorithm along with our fitness equations design (Equations~\ref{eq:fitness_argmin} and~\ref{eq:fitness_graph_formulation}). A particular case is the result of the Driving 0p video in which the PSO parameter setting have a negative improvement over the manual tuning. In this experiment, the PSO algorithm selected a higher speed-up rate for the non-semantic segments in comparison to our best manual tuning, leading our method to select frames with large temporal distance.

\begin{figure}[t!]
	\centering
	\includegraphics[width=0.75\textwidth]{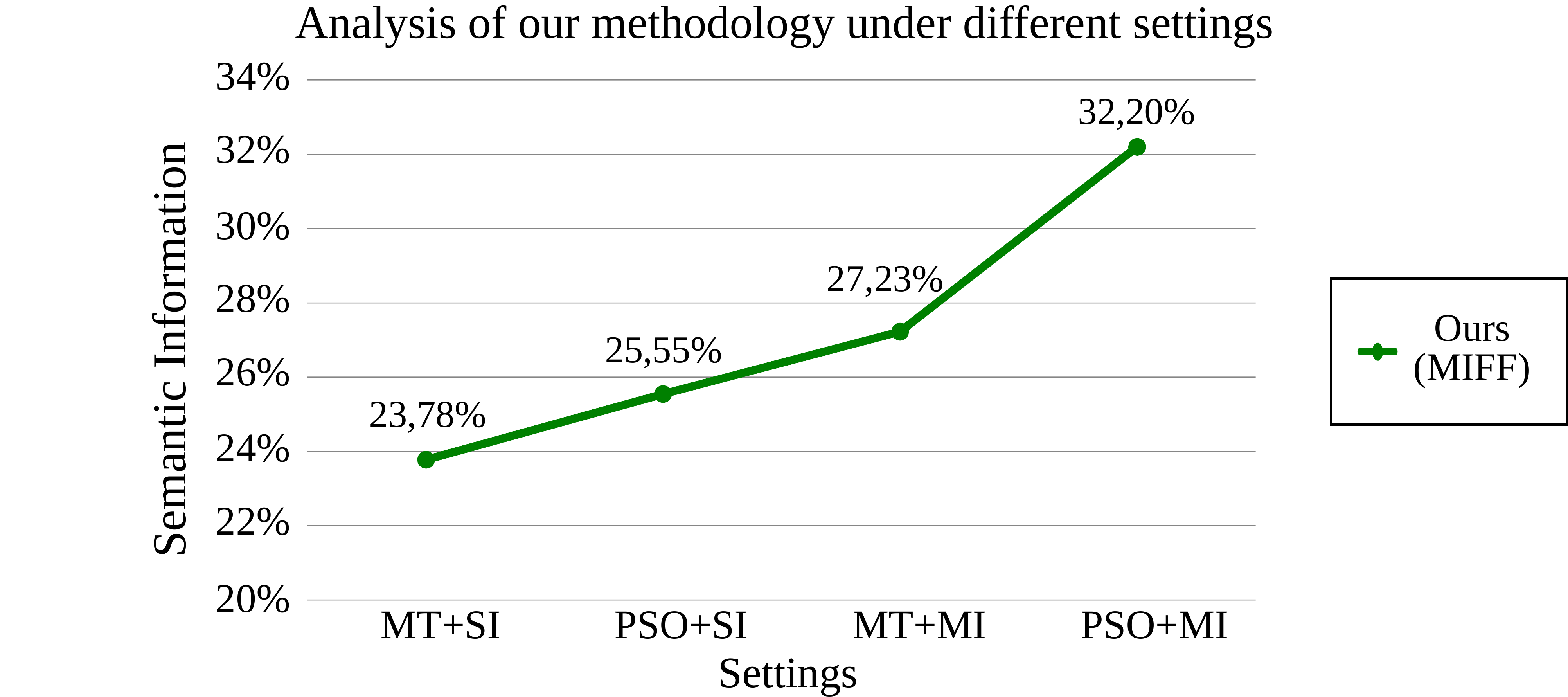}
	\caption{Average semantic information retained under different parameter settings. We tested a combination of the Single-Importance (SI) or Multi-Importance approaches (MI) with the best Manual Tuning (MT) or Particle Swarm Optimization (PSO). The values are related to the maximum semantic information possible to obtain in the fast-forward video. The combination PSO+MI leads to the highest amount of semantics in the final video.}
	\label{fig:Semantic_Analysis_of_the_Technique_Evolution}
\end{figure}

Figure~\ref{fig:Semantic_Analysis_of_the_Technique_Evolution} shows a more detailed performance assessment, considering both the parameter setting and frame selection algorithms, namely: Single-Importance (SI) or Multi-Importance approaches (MI) with the best Manual Tuning (MT) or Particle Swarm Optimization (PSO). The values are related to the maximum amount of semantic information possible in a fast-forward video given the required speed-up. One can clearly see that MI approach contributes for the methodology to keep more semantic information compared to SI approach. Further, the results indicate that using the Multi-importance approach jointly with the PSO parameters setting produces output videos with an average of $9$ percentage points with more semantic information than the MT+SI.

\paragraph{Video stabilization}

\begin{figure}[t!]
	\centering
	\includegraphics[width=0.95\textwidth]{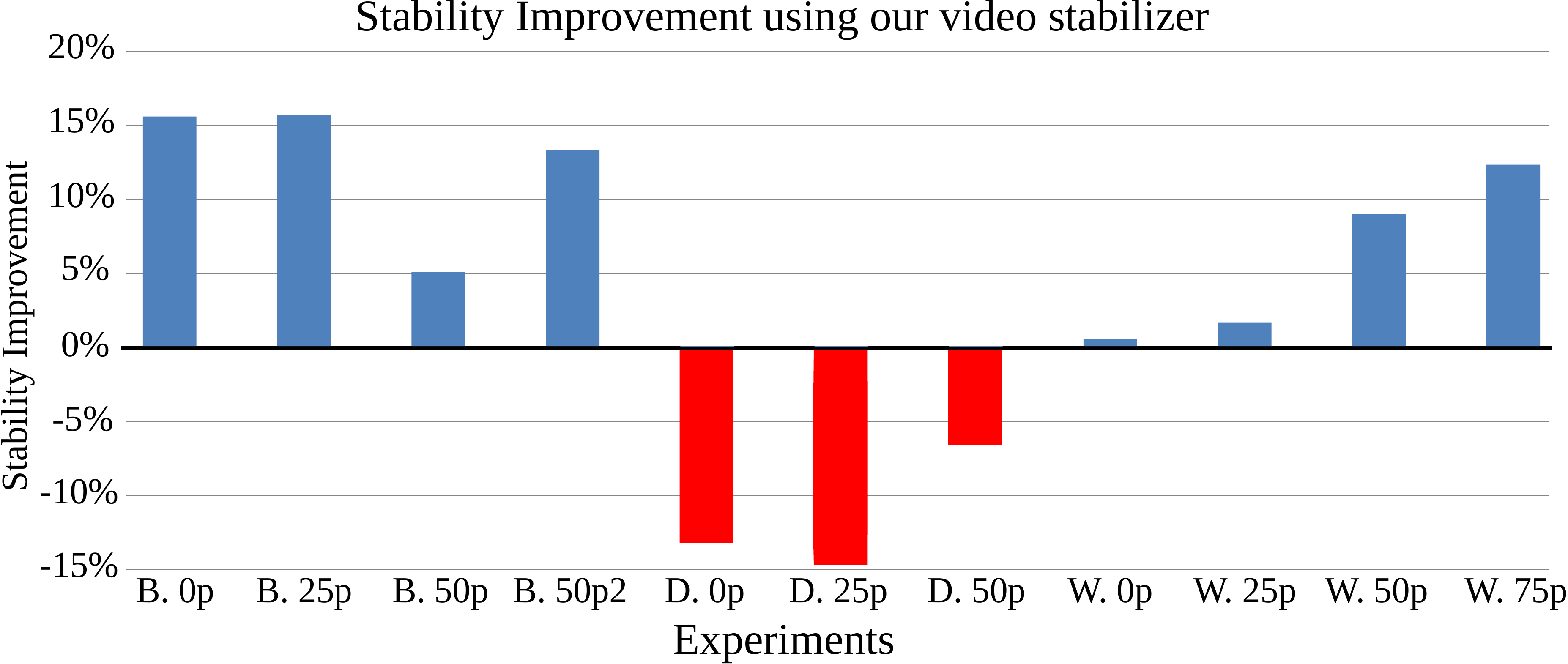}
	\caption{Stabilization improvement over the semantic fast-forward step. An improvement of $100\%$ indicates that the output video is as stable as the original one. The W., D., and B. stand for Walking, Driving, and Biking experiments, respectively.}
	\label{fig:VideoStabilization_Evaluation}
\end{figure}

We compared the instability of the videos generated before and after the semantic fast-forwarding. Our stabilization step achieved the best results for all videos in the Unlabeled Dataset. Figure~\ref{fig:VideoStabilization_Evaluation} depicts the results for the Semantic Dataset. An improvement of $ 100\% $ in the instability indicates that the output video is as stable as the original one.

It can be seen that in most cases, our stabilization step presents an improvement over the semantic fast-forwarding step. The Driving experiments are failure cases of our stabilization approach. The high speed of motion of a car causes small scene overlaps between the fast-forward video frames. Thus, due to the features mismatches, the target homography planes are erroneously computed, leading the video to present unstable transitions. The average stability improvement in the Semantic Dataset is $ 3.48\% $, however, excluding the Driving experiments, the value increases to $ 9.09\% $.

A more detailed performance assessment of stabilizing fast-forward egocentric videos was performed by comparing our stabilization method with the work of Joshi~\etal~\cite{Joshi2015} (MSH), which is a smoothed homography frame-to-frame transformation. We create a video using the frames selected by the MSH frame sampling step. Then, we execute our stabilization step on this video. To evaluate the smoothness, we compare the values of the instability index of this stabilized video with the MSH video.

The average of the instability values over all experiments of the videos stabilized by our technique was equal to $ 35.04 $, facing $ 34.04 $ of the ones stabilized by MSH stabilizer. However, our methodology has not been designed to perform well with larger movements, like driving. Then, considering the `Driving' sequences as outliers samples and not including them in the average computation, our stabilizer presents an average instability of $ 32.30 $ against $ 32.54 $ of the MSH stabilizer. Further, in Joshi~\etal's work~\cite{Joshi2015}, they stated that their frame selection is optimal. Therefore, our video stabilizer outperforms theirs in the best set of frames.

\paragraph{Method comparison for motion estimation}

A good motion estimation is crucial for selecting a set of frames that reduces the shakiness on the fast-forward videos. Some adaptive frame selection algorithms adopt the Focus of Expansion (FOE) from the Sparse Optical Flow to make the motion estimation~\cite{Poleg2015, Ramos2016, Silva2016, Halperin2017}. We investigated the influence of changing the estimator in our methodology comparing four different estimators: Phase Correlation, FlowNetSimple, FlowNetCorr, Sparse Optical Flow (SOF). The Phase Correlation is a method that measures the pixel displacement between two images using the magnitude of the Fourier Transform~\cite{Tekalp95}. The FlowNetSimple and the FlowNetCorr are two network architectures designed to estimate optical flow using Convolutional Neural Networks~(CNN)~\cite{Dosovitskiy2015}. The SOF is a cumulative optical flow technique based on Lukas-Kanade method, it was proposed to long-term temporal segmentation of Egocentric Vision~\cite{Poleg2014}. 

\begin{figure}[t!]
	\centering
	\includegraphics[width=0.99\textwidth]{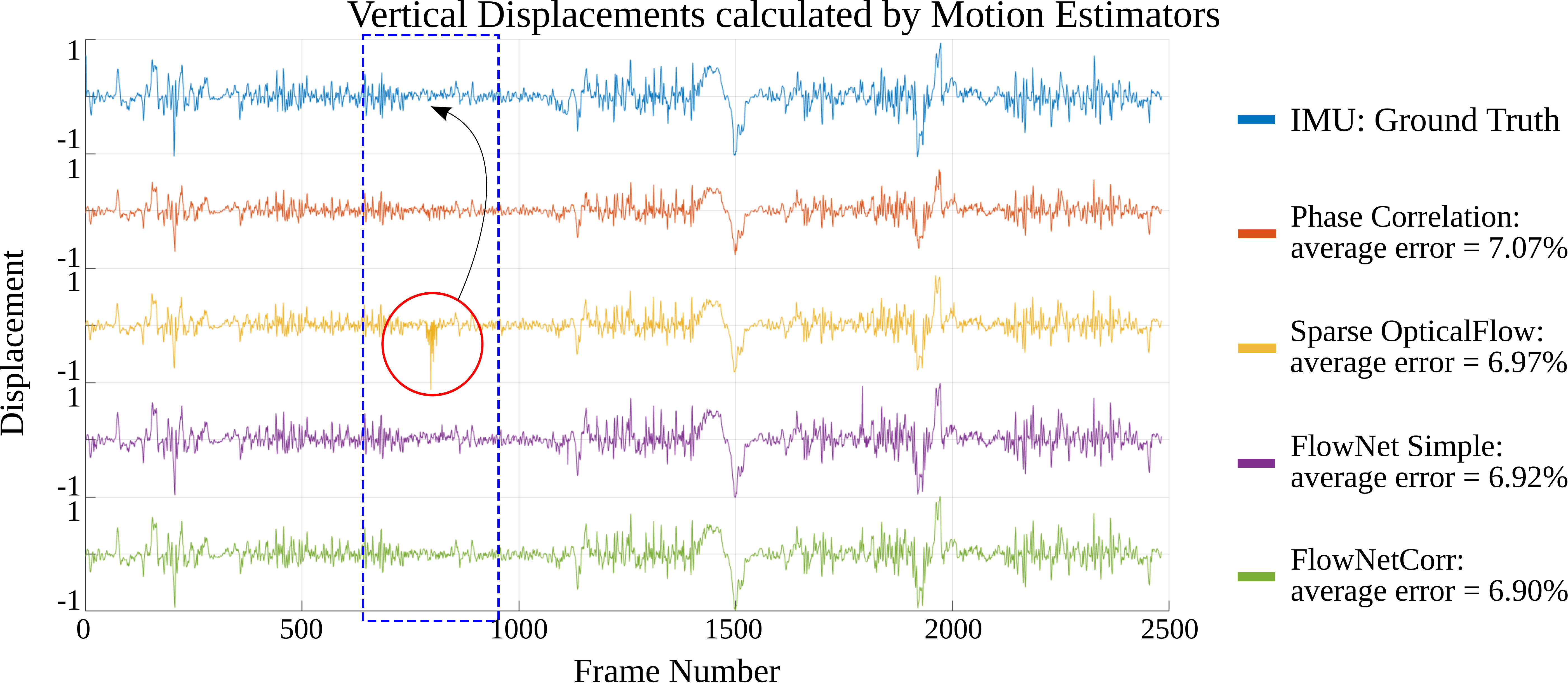}
	\caption{Vertical displacements calculated by different motion estimators in one of our videos. The circled area indicates a failure case in the Sparse Optical Flow motion estimation that does not occur in the other estimators. The average Root Mean Squared Error indicates FlowNetCorr as the best estimator.}
	\label{fig:Vertical_Displacements}
\end{figure}

To evaluate their estimations, we recorded five videos covering actions such as walking, going up the stairs, turning from left to right, from up to down, and vice versa. All videos were recorded using an Inertial Measurement Unit (IMU). After applying the estimators, we compare their measurements with the IMU records. Figure~\ref{fig:Vertical_Displacements} shows a vertical displacements plot of one of the videos, and the average of the Root Mean Squared Error (RMSE) for each method over all videos. 
We can see that the Sparse Optical Flow is one of the estimator with the highest average RMSE, meaning its motion estimation is not accurate. Once the FlowNetCorr is the motion estimator with the lowest average, we analyze the effect of using it in our methodology instead of the Sparse Optical Flow. 

\begin{table}[t!]
	\centering
	\caption{Method Comparison between Sparse Optical Flow (SOF) and FlowNetCorr (FNC).}
	\label{tab:OF_analysis}
	\footnotesize{
		\begin{tabular}{@{}lrrrrrrrr@{}}
			\toprule
			& \multicolumn{2}{c}{Semantic Information$^{1}$} & &  \multicolumn{2}{c}{Instability$^{2}$}  & &  \multicolumn{2}{c}{Speed-up$^{3}$}   \\ 			
			& \textbf{SOF}    & \textbf{FNC}   & &  \textbf{SOF} & \textbf{FNC} & &  \textbf{SOF} & \textbf{FNC} \\ \cmidrule(l){2-3} \cmidrule(l){5-6} \cmidrule(l){8-9}
			
			B.0p    & \textbf{20.60\%}  & 20.53\%    		& &  24.84   & \textbf{24.80}  & &  12.56   & \textbf{11.47}  \\
			B.25p   & \textbf{41.64\%}  & 39.38\%  			& &  \textbf{44.84}   & 46.41  & &  \textbf{10.01}   & 9.88  \\
			B.50p   & \textbf{33.67\%}  & 24.52\%  			& &  31.18   & \textbf{30.36}  & &  10.05   & \textbf{10.01}  \\
			B.50p 2 & 25.42\%       	& \textbf{25.83\%} 	& &  27.11   & \textbf{26.92}  & &  9.87   & \textbf{10.01}   \\
			D.0p    & \textbf{31.81\%}  & 29.69\%     		& &  48.31   & \textbf{47.79}  & &  13.83   & \textbf{12.37}  \\
			D.25p   & \textbf{46.38\%}  & 45.69\%    		& &  \textbf{34.49}   & 38.52  & &  10.07   & \textbf{10.00}  \\
			D.50p   & \textbf{18.56\%}  & 17.78\%    		& &  40.24   & \textbf{39.50}  & &  \textbf{10.02}   & 9.89  \\
			W.0p    & \textbf{15.20\%}  & \textbf{15.20\%} 	& &  \textbf{35.18}   & 35.43  & &  \textbf{10.00}   & \textbf{10.00}   \\
			W.25p   & 39.55\%   		& \textbf{51.20\%} 	& &  32.47   & \textbf{31.26}  & &  9.84   & \textbf{10.03}   \\
			W.50p   & 26.35\%   		& \textbf{37.11\%} 	& &  34.57   & \textbf{34.11}  & &  \textbf{9.94}   & 9.82  \\
			W.75p   & 47.09\%			& \textbf{47.29\%}  & &  34.38   & \textbf{34.37}  & &  8.31    & \textbf{9.23}   \\ \cmidrule(l){2-3} \cmidrule(l){5-6} \cmidrule(l){8-9}
			
			\textbf{Mean}& 31.48\% & \textbf{32.20\%}  & & \textbf{35.23}  & 35.41 & &  10.41   & \textbf{10.25}   \\
			& \multicolumn{2}{c}{\footnotesize{\textit{$^{1}$Higher is better}}}  & & \multicolumn{2}{c}{\footnotesize{\textit{$^{2}$Lower is better}}} & &  \multicolumn{2}{c}{\footnotesize{\textit{$^{3}$Better close to 10}}} \\ 	 \bottomrule
		\end{tabular}
	}
\end{table} 

The methodology has a better performance when using the FlowNetCorr, since it decreases the frequency of wrong measurements. For example, if the recorder suddenly turns the camera and the estimated movement is low, it may lead the methodology to treat the movement as stationary. Table~\ref{tab:OF_analysis} shows that changing the motion estimator indeed improves our method considering the Semantic and Speed-up metric, while still being closer to the best result for instability metric.

\subsection{Qualitative analysis}

Numerous questions arise when the word ``semantic'' shows up in a work: ``\textit{What is semantic information?}'', ``\textit{How do you define it?}'' and ``\textit{Why do you consider something as semantic?}''. Next we clarify what is meant by semantic in this work. 

In Section~\ref{sec:experiments_evaluation_metrics}, due to the need of establishing a ground truth for comparison and exhibition purposes, we perform the experiments using a defined semantic: face detection for videos with slow movements and pedestrian for the others. Obviously, semantics is much more than faces or pedestrians -- it may be considered as everything that visually attracts the user's attention. In this Section, we show how to classify the frames using the proposed approach to classifying semantic contents based on the user's preference from web video statistics.

\paragraph{Dataset} 

To mine the information about the universal interest, we work with the available statistic data of YouTube videos. Further, the images composing the selected videos are used in the training process to learn how to identify a ``Cool'' frame. We fine-tune and test the CNN on a dataset composed of videos collected from the YouTube8M~\cite{Abu-El-Haija2016}. Because our focus is egocentric videos, we filter the list using the keyword ``GoPro''. We rank the videos according to the score $C = \frac{views}{(dislikes/likes)}$ and select the $150$-top ranked videos to compose the \textit{Cool} class.

Analyzing the selected videos, we found that most of them were related to radical sports and pleasant landscapes. Therefore, to compose the ``Not Cool'' class, we manually selected $150$ videos from the labels with the opposite concept, such as ``Home Video'', ``Mobile Home'', ``Office'', and ``House''. Finally, after removing the intros, edition effects and blurred frames, the final dataset contains a total $ 940{,}030 $ labeled images.

\paragraph{Training}

We consider the estimation of the semantic content in a frame similar to the scene recognition problem, \ie, we analyze the entire frame. Therefore, we started the training from a VGG16 model trained on MIT Places205~\cite{Zhou2014} and fine-tuned it in our dataset. The training step used $80\%$ of the dataset, and $20\%$ of this data was used for validation purposes. After running a random search to tune the learning parameters, we set $1\times10^{-6}$ for \textit{base\_lr} and $5\times10^{-4}$ for \textit{weight\_decay}. The final network's accuracy was $98.03\%$.

\paragraph{Results}
\begin{figure}[h]
	\begin{center}
		\includegraphics[width=0.95\textwidth]{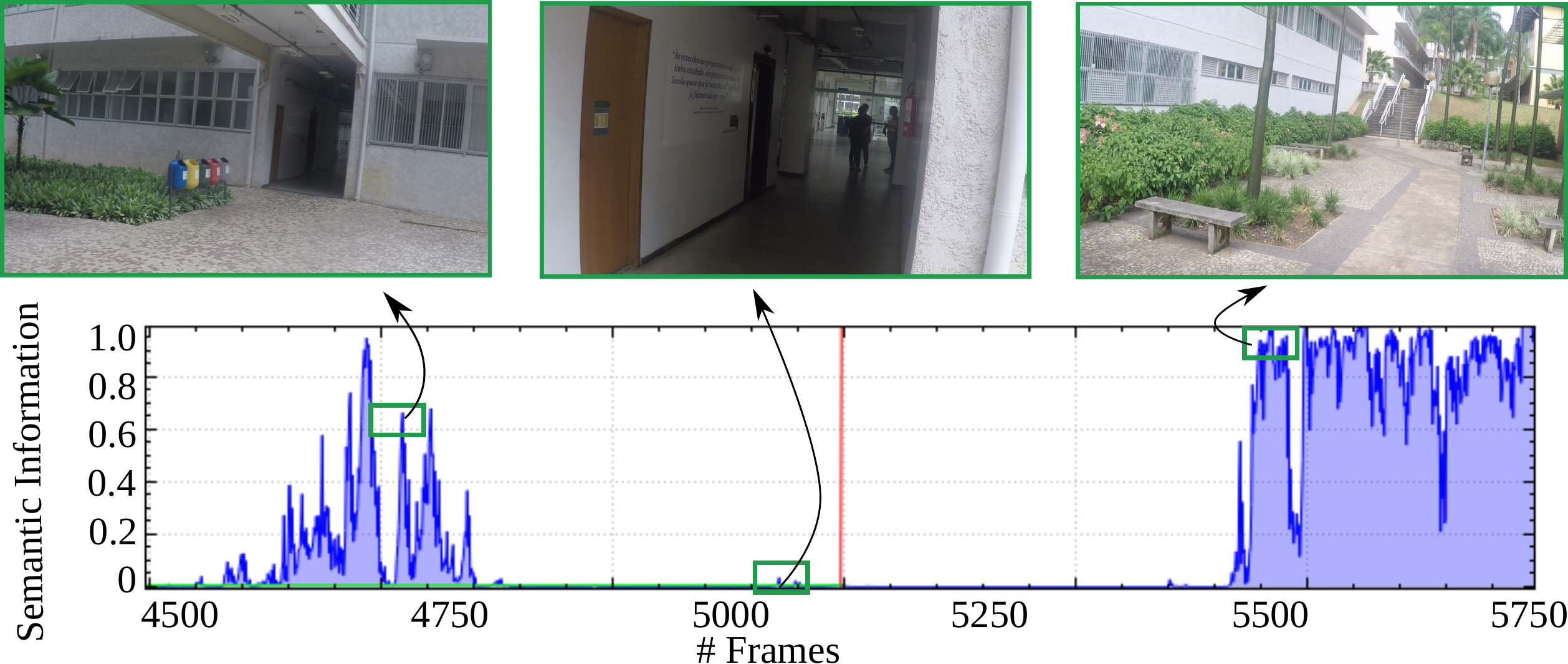}
		\caption{Semantic Profile curve of the CoolNet for every frame of a sample video. The left image depicts an inside garden, with its medium score. The central image is a building hall, that the CoolNet does not consider containing large semantic content. The right image is a garden with a outdoor view, for which CoolNet gives the highest scores.} 
		\label{fig:Cool_net}
	\end{center}
\end{figure}

Since most of the ``Cool'' images in our dataset are related to radical sports and beautiful landscapes, the network classifies with high score frames with nature-related elements, \eg, forest and gardens. Uniform scene frames, like indoor looking images, walls, and offices, yield to a low rating. Figure~\ref{fig:Cool_net} depicts network scores related to different scenes. In the left image, when the wearer passes through an inside garden, the network attributes an average rating. In the center image, the wearer is walking inside a building hall, which the net considers unattractive. In the right image, the wearer goes to an outside area composed of many trees and gardens, which are highly rated by the net.

\paragraph{Semantic combination}

Although the \textit{CoolNet} incorporates user's preferences to estimate the semantic in each frame, it could be not enough to cover all considered semantics. We address this issue by combining semantic extractors, making a linear combination of their output. In this case, the output is a fast-forward video emphasizing segments which have either faces or beautiful landscapes, for example. Additionally, since the score for each frame is given by a linear combination, we can set which extractor has more influence. Then, we can make it individual, analyzing the user's browsing behavior, website or social network profiles, similar to recommendation systems. The reader is referred to our supplementary video for a visual result of this combination.

\section{Conclusions}
\label{sec:conclusions_and_future_works}

In this work, we proposed the Multi-Importance Fast-Forward~(MIFF), a fully automatic technique to produce shorter versions of egocentric videos given more emphasis to their semantic content. To make it automatic, we use a parameter setting via Particle Swarm Optimization algorithm. Contrasting with previous semantic fast-forward that estimate just one speed-up for the semantic portion, we propose a Multi-Importance approach to emphasize proportionally to the relevance of the segment. These new contributions enabled our methodology to keep over $3$ times more semantic content than the state-of-the-art method. 

As expected, lending more emphasis to the semantic segments makes the non-semantic ones run faster and jerkier. To overcome this problem, we applied and discussed the need of a Video-Stabilizer specific for fast-forward egocentric video. Regarding the dilemma of semantic definition, we proposed a methodology -- be it general or customized -- to learn the semantics from the user preferences. \adding{The customized aspect can be achieved during the semantic definition by using a set of classifiers covering all user intentions or training the \textit{CoolNet} in the data gathered from the user Youtube profile instead of general statistics.} To validate the presented methodology, we ran quantitative experiments, based on fixed and specific semantics, and qualitative experiments, using the general semantic approach proposed here. Throughout the discussion, we presented the drawbacks of video stabilization when camera motion is high, like recording from the inside of a moving car. Identifying which segments the method is able to stabilize and how to deal with fast movements will be the focus of future investigation.

\paragraph{\textbf{Acknowledgments}}
The authors would like to thank the agencies CAPES, CNPq, FAPEMIG, and Petrobras for funding different parts of this work.

\section*{References}

\bibliography{JVCIR_2018}

\end{document}